\documentclass[letterpaper, 10 pt, conference]{ieeeconf}  

\IEEEoverridecommandlockouts                              

\usepackage{hyperref}
\usepackage{url}
\usepackage{graphics} 
\usepackage{times} 
\usepackage{booktabs}       
\usepackage{amsfonts}       
\usepackage{nicefrac}       
\usepackage{microtype}      
\usepackage{algorithm,multirow,xcolor}
\usepackage{algorithmicx}
\usepackage{algpseudocode}
\usepackage{mathtools}
\usepackage{graphicx}
\usepackage{cases}
\usepackage{array}
\usepackage{color}
\usepackage{float}
\usepackage{cite}
\usepackage[skip=6pt]{caption}

\usepackage{wrapfig}
\usepackage{subfigure}
\usepackage{amsmath}
\usepackage{amsthm, amssymb}
\theoremstyle{definition}

\usepackage{soul}
\usepackage{epstopdf}


\title{\LARGE \bf
	Reinforced Potential Field for Multi-Robot Motion Planning in \\ Cluttered Environments
}

\author{Dengyu Zhang, Xinyu Zhang,   Zheng Zhang, 
 Bo Zhu, and Qingrui Zhang 
\thanks{All authors are with the School of Aeronautics and Astronautics, Sun Yat-sen University, Shenzhen 518107, P.R. China (\{ zhangdy56, zhangxy385, zhangzh363\}@mail2.sysu.edu.cn, \{zhubo5, zhangqr9\}@mail.sysu.edu.cn; Corresponding author: Qingrui Zhang)}
}

\begin{document}
	
	\maketitle
	\thispagestyle{empty}
	\pagestyle{empty}

	\begin{abstract}
 Motion planning is challenging for multiple robots in cluttered environments without communication, especially in view of real-time efficiency, motion safety, distributed computation, and trajectory optimality, \emph{etc}. In this paper, a reinforced potential field method is developed for distributed multi-robot motion planning, which is a synthesized design of reinforcement learning and artificial potential fields. An observation embedding with a self-attention mechanism is presented to model the robot-robot and robot-environment interactions. A soft wall-following rule is developed to improve the trajectory smoothness. Our method belongs to reactive planning, but environment properties are implicitly encoded. The total amount of robots in our method can be scaled up to any number.  The performance improvement over a vanilla APF and RL method has been demonstrated via numerical simulations. Experiments are also performed using quadrotors to further illustrate the competence of our method.
	\end{abstract}

	
	\section{INTRODUCTION}

With recent advances in automatic technologies and artificial intelligence, robots are envisioned to run efficiently, reliably, and safely with little human intervention in a myriad of applications, \emph{e.g.,} package delivery, planetary exploration, supply chain automation, and search-and-rescue operation \emph{etc.} \cite{wurm2008IROS, Mathew2015TASE,bai2018RAS}. The burgeoning applications in diverse tasks put great demands on robot autonomy to support safe and intelligent operations in complex environments. Motion planning is one of the fundamental modules of robot autonomy, which computes dynamically-feasible trajectories for robots to reach their respective goal regions \cite{mohanan2018RAS}. Yet, motion planning is a challenging task, especially for multiple robots with no communication in cluttered environments \cite{snape2010IIRS, Zhao2018ACC}. Finding a safe and smooth motion towards a target is demanding in a congested environment with numerous obstacles.  A robot also needs to comply with motions of other robots in the shared space, while communications among robots are scarcely available \cite{chen2017IROS,kretzschmar2016IJRR}.
The decision difficulty, planning cost, and collision risk will all grow exponentially with the amount of robots and obstacles in cluttered environments. 

Most methods of motion planning in multi-robot cluttered environments fall into two categories, namely proactive and reactive algorithms. In proactive planning, robots make active decisions based on current and forthcoming states of the environment \cite{Sandeep2019IJRR,Alonso2018TRO,Truong2017TASE}. Hence, sophisticated models are prerequisite for the prediction of environment states in the near future, interactive information among neighbouring robots, or intentions of other intelligent agents \cite{Luber2012IROS}. Most proactive planning algorithms are based on online optimization, \emph{e.g.}, dynamic window approach (DWA), model predictive control (MPC) \cite{zhu2019IRAL,soria2021Nature}, \emph{etc}. The computation of those optimization-based algorithms is cumbersome for lightweight platforms in cluttered environments, \emph{e.g.}, UAVs and UGVs, etc. Convergence is not ensured in short-time windows. Communication is required among robots to share prediction results. 

 \begin{figure}[tbp]
     \centering
     \includegraphics[width=\linewidth]{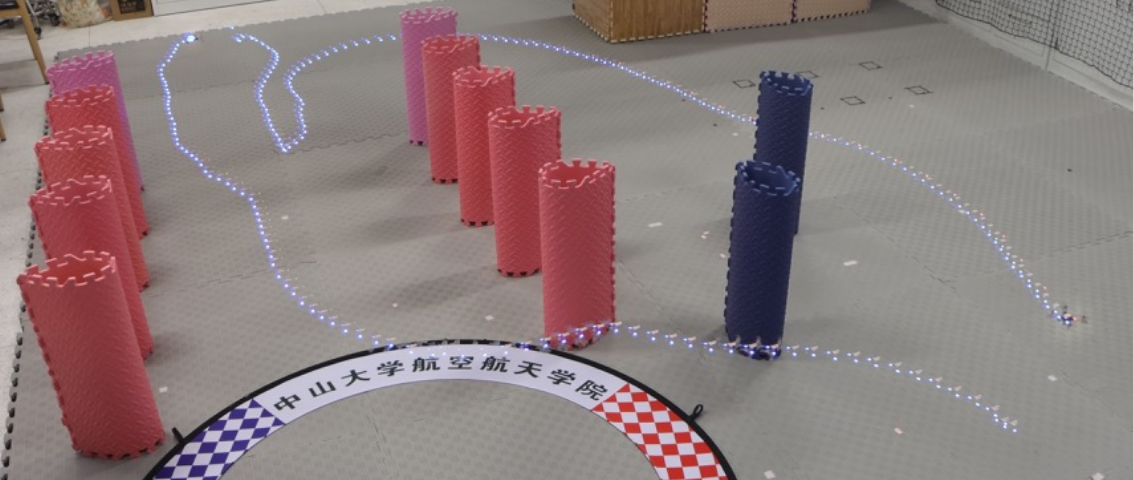}
     \caption{Multiple Quadrotors flying in a cluttered environment by the reinforced potential field algorithm.}
     \label{fig:APFStuck}
 \end{figure}

In reactive planning, robots determine their motion directly based on the instantaneous measurements from their proprioceptive and  exteroceptive sensors \cite{arslan2019IJRR,cole2018ITR}. Robot motion is generated in real-time as a function of a certain vector field or virtual forces, \emph{e.g.}, guiding vector field \cite{Yao2021TRO}, Dipole-like field \cite{panagou2014ICRA}, and artificial potential field (APF) \cite{ZhangZheng2022}, \emph{etc.} Although solutions for reactive planning are not optimal in general, they are mostly computationally faster than the optimization-based proactive methods.  Communication among robots is not necessary for reactive planning, which is an advantage for distributed scenarios. Hence, reactive planning is suitable for lightweight platforms with limited resources. However, non-smooth and oscillatory trajectories are often generated by reactive planning. Conventional reactive planning methods would experience the deadlock issue in multi-robot cluttered environments. 


In this paper, a reinforced potential field (RPF) method is proposed for distributed multi-robot motion planning in cluttered environments. The proposed method belongs to reactive planning {methods}. In the proposed design, reinforcement learning (RL) is introduced to implicitly encode the diverse impact of surrounding environments and capture motions of other robots in the shared space. The dynamic changes in surroundings will be feedback into an APF module, so robots can adjust their motion in time. An observation embedding with self-attention mechanism is employed to characterize both the robot-environment interactions and the interactive information among robots, thereby inferring the relative importance of neighboring agents or obstacles. A soft wall-following rule is designed to smooth the APF outputs. The efficiency of the overall algorithm is evaluated extensively using numerical simulations, and also validated through experiments with quadrotors. In summary, the main contributions of this paper are three fold:
 \begin{enumerate}
     \item A distributed reactive planning framework is developed for multiple robots in cluttered environments. The proposed design integrates RL into conventional reactive planning methods. With the inference capability of RL, the proposed design is able to make active responses to dynamic change of surrounding environments. 
     \item An observation embedding with self-attention mechanism is {introduced to our algorithm}, which is able to extract the hidden features of the  environments.
     \item A soft wall-following rule is presented, which could improve the smoothness of the planned trajectory.
 \end{enumerate}

    The rest of this paper is organized as follows. In Section \ref{sec:Related}, the related works are summarized. Preliminaries are provided in Section \ref{sec:preli}. Section \ref{sec:alg} presents the implementation details of the proposed algorithm. Numerical simulations and experimental results are given in Section \ref{sec:exp}. Conclusion remarks are summarized in Section \ref{sec:Conclusion}.
	
	\section{Related Works}
 \label{sec:Related}
    In model-based proactive planning, forthcoming states of the environment are used in the current decision-making. Several methods have been proposed, such as  velocity obstacles (VOs) \cite{tan2020IRS}, decentralized MPC \cite{Tallamraju2018SSRR} and sequential MPC \cite{Aoki2022ITSC}. VO-based methods make predictions in light of the current velocity, where uncertainties are handled by extending bounding volumes \cite{tan2020IRS}. The VO method was modified in an optimal reciprocal collision avoidance (ORCA) approach to improve the trajectory  smoothness \cite{snape2010IIRS}. Based on VO or ORCA, both decentralized and sequential MPC algorithms are studied for multi-robot motion planning \cite{Alonso2018TRO}. 
Proactive methods can achieve safe collision avoidance with good performance. However, communication is necessary for intelligent robots in MPC to share their planned decisions in the forthcoming horizon. The smooth and safe planning of MPC is obtained at the cost of high computation.  When the amount of robots and obstacles grows, these methods may be neither available nor reliable for small-scale robots with limited resources.

Reactive planning methods are designed mainly based on Artificial Potential Fields (APFs). In APFs, two types of virtual forces are defined to regulate robots' motion, namely attractive force for target reaching and  repulsive force for collision avoidance \cite{Khatib1986APF}.  APF-based methods are capable of offering safe robot navigation in sparse stationary environments \cite{ge2002AR,mohanan2018RAS}. However, APF-based methods suffer from the local minima issue, so no assurance is guaranteed for global convergence to a target location \cite{mohanan2018RAS}. In \cite{arslan2019IJRR}, the local minima issue was addressed for convex cluttered environments with sufficiently separated obstacles. The issue is still open for general cluttered environments.  An alternative reactive planning is based on vector fields (VFs) that could avoid the local minima issue \cite{Yao2021TRO}.
In \cite{panagou2014ICRA}, a Dipole-like vector field is designed for multi-robot motion planning. Global convergence is ensured for environments with sufficiently separated obstacles \cite{panagou2014ICRA}. However, VF-based methods assume the  environment is fully known a priori. 

    In learning-based methods, collision-free planning policies are learned mainly by reinforcement learning  (RL) through the maximization of certain return functions using data samples\cite{kretzschmar2016IJRR,riviere2020IRAL}.  In \cite{chen2017IROS}, a deep value learning method is used to generate human-like navigation for robots in crowded dynamic environments. In \cite{chen2019ICRA}, the deep value learning approach is modified by adding a socially attentive network that models crowd-robot interactions. In \cite{Tingxiang2020IJRR}, a sensor-level decentralized collision-avoidance policy is learned using proximal policy optimization (PPO) method, which has an end-to-end feature. Vanilla learning-based methods always suffer from low data efficiency and poor generalization. Hence, a synthesized design of learning-based and force-based methods is proposed for distributed motion planning in dynamic dense environments \cite{Semnani2020IRAL}. In our previous works \cite{ZhangZheng2022}, artificial potential field is also integrated into the Dueling Double Deep Q Network (D3QN) to boost the policy learning for the multi-robot cooperative hunting problem.  In this paper, we will develop a new synthesized design that explicitly models robot-environment interactions.  Future environment states are used to further improve the algorithm's performance.

	\section{Preliminaries}
 \label{sec:preli}
	\subsection{Problem Formulation}\label{sec:problem_formulation}
In this paper, the motion planning problem is considered for $N$ robots navigating to their preassigned goals in the same environment with obstacles ($N\geq1$). The total set of robots is defined to be $\mathcal{V}=\left\{1,\;2\,\ldots,N\right\}$. All robots share the same space with different goal positions, so there might be conflicts for robots to directly move towards their respective goals.  The environment is partially observable for each robot. All robots move at a desired speed $v$. They can only change their heading angles for both reaching goals and avoiding collision \footnote{Note that the proposed framework can be easily extended to the case with a changing desired speed.}. Such a task is challenging, as robot only has one input to change its motion for two tasks in cluttered multi-robot environments.  This paper is the upper-level motion planning, so a well-designed inner-loop controller is assumed to be implemented on the robot.  Hence, a first-order point-mass robot is considered for a robot $\forall$ $i\in\mathcal{V}$ as below. 
    \begin{equation}
        \dot{\boldsymbol{p}}_i = \boldsymbol{v}_i 
        \label{eq:SysDyn}
    \end{equation}
where $\boldsymbol{p}_i\in\mathbb{R}^m$ is the robot position vector with $m\geq2$, and $\boldsymbol{v}_i\in\mathbb{R}^m$ is the velocity vector.

    \begin{figure}[tbp]
        \centering        
        \includegraphics[width=0.95\linewidth]{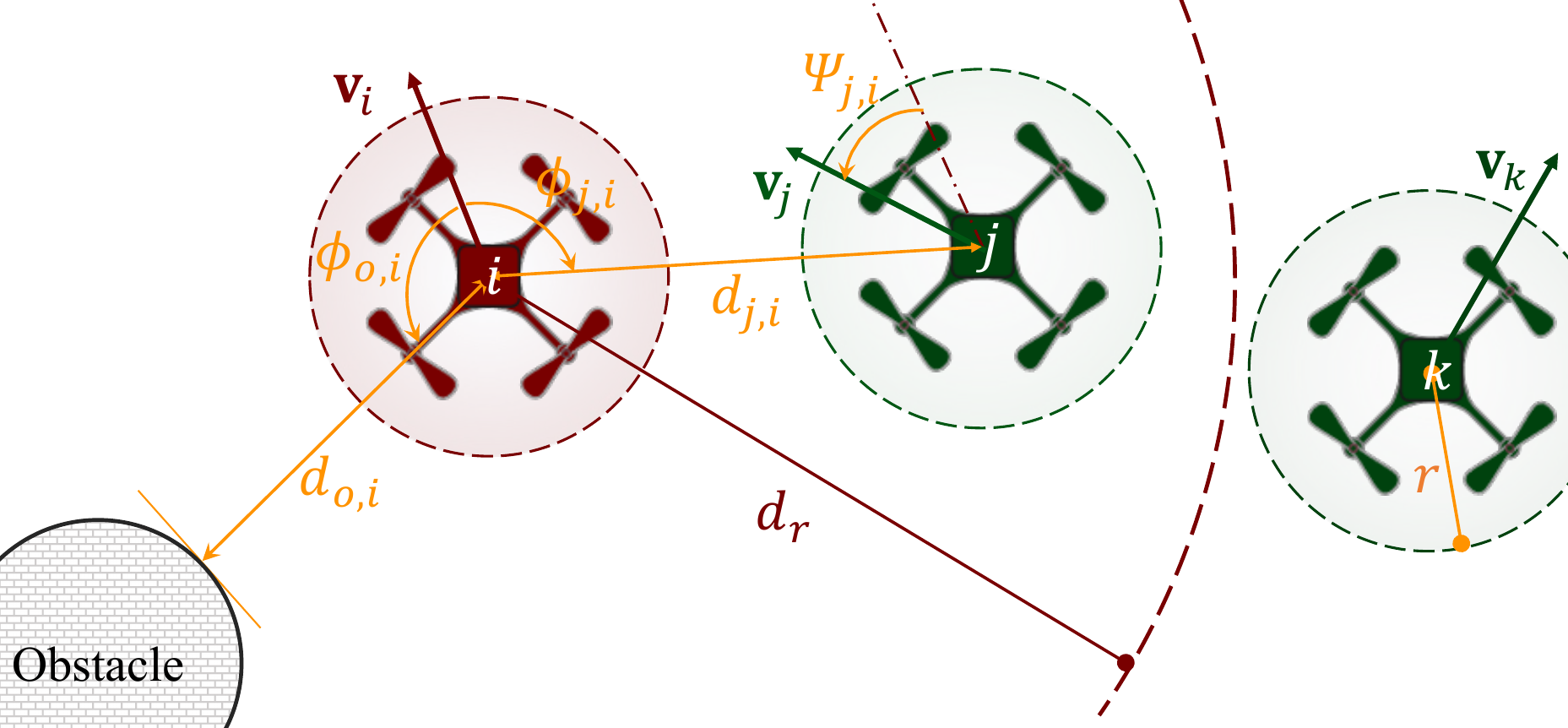}
        \caption{Problem background (All robots are homogeneous with the same safe radius dented by $r$. A robot $i$ is able to detect obstacles and neighbouring robots in a range of $d_r$. Necessary exteroceptive sensors are mounted on the robot $i$ to obtain surrounding information, such as the obstacle distance $d_{o, i}$, the azimuth angle of an obstacle, the distance of a neighbouring robot $d_{j, i}$,  azimuth angle of a neighbouring robot $\phi_{j,i}$, and the relative heading angle $\psi_{j,i}$ between robot $i$ and its neighbour $j$.).}
        \label{fig:problem_formulation}
    \end{figure}

As shown in Fig. \ref{fig:problem_formulation}, all robots of interest in the environment are homogeneous, which has a safe radius of $r$.  Let $d_{o, i}$  be the distance from the robot $i$ to the surface of an obstacle, and $d_{j, i}= \|\boldsymbol{p}_j-\boldsymbol{p}_i\|$ be the distance between the robots $j$ and $i$ as in Fig. \ref{fig:problem_formulation}. Hence, collisions will occur if $d_{o, i}<r$ or $d_{j,i}<2r$. We use $\phi_{o,i}$ to denote the azimuth angle of an obstacle in the local frame of the robot $i$ as shown in Fig. \ref{fig:problem_formulation}. Similarly, $\phi_{j,i}$ is the azimuth angle of a robot $j$ in the local frame of the robot $i$. The relative heading angle of a robot $j$ to robot $i$ is specified by $\psi_{j,i}$.  Assume that exteroceptive sensors are available for each robot $i$ to observe the environment, \emph{e.g.}, LiDAR or visual sensors.  The detection or sensing range is represented by $d_r$ for each robot as illustrated in Fig. \ref{fig:problem_formulation}. Hence, based on the exteroceptive sensors, an robot $i$ is able to obtain all necessary environment states in its detection range for planning, including obstacle information \{$d_{o, i}$, $\phi_{o,i}$\} and neighouring robot information \{$d_{j,i}$, $\phi_{j,i}$, $\psi_{j,i}$\}. The goal position for an robot $i$ is represented by $\boldsymbol{p}_g$, so $d_{g,i}=\|\boldsymbol{p}_{g,i}-\boldsymbol{p}_i\|$. In summary, the objective of this paper is  formulated as 
\begin{equation} \label{eq:Obj}
\left\{\begin{array}{l}
     \lim_{t\to\infty} d_{g,i}(t)<r \text{ with }  d_{g,i}(0)>0   \\
     d_{j,i}(t)>2r \quad \forall\; j\in\mathcal{V}, \text{ } t>0, \text{ \& } j\neq i \\
     \text{No collision with any obstacle}
\end{array}
\right.
    \forall\; i\in \mathcal{V}
\end{equation}

 

 \subsection{Reinforcement Learning}\label{sec:RL}
 In RL, state transitions of environments are formulated as a Markov Decision Process (MDP) that is denoted by a tuple
$(\mathcal{S},\mathcal{A},\mathcal{P},{R},\gamma)$, where
	\begin{itemize}
		\item $\mathcal S$ is the state space with $s_t \in \mathcal S$ denoting the environment state at the timestep $t$.
		\item $\mathcal A$ is the action space with $a_t \in \mathcal A$ being the action executed at the timestep  $t$.
		\item $\mathcal{P}$ is the state-transition probability function.
		\item ${R}$ is the reward function; 
		\item $\gamma \in \left(0,1\right)$ is a discount factor balancing the instant reward with future reward.
	\end{itemize}
Let $\pi(a|s)$ be the stochastic policy in RL with $\pi(a|s) : \mathcal{S}\times \mathcal{A} \to \left[0\; 1\right]$. The expectation of the accumulated reward $R_t$ is 
 \begin{equation}
     \mathbb{E} [\mathcal{R}_t]= \mathbb{E}_{a\sim\pi, s\sim\mathcal{P}}[\sum_{t=k}^T\gamma^{t-k} R_t] \label{eq:ReturnRt}
 \end{equation}
  where $a_t\sim\pi(a_t|s_t)$, $s_{t+1}\sim \mathcal{P}(s_{t+1} | s_t,a_t)$, and $T$ is the task horizon.
 The objective of RL is to learn an optimal policy $\pi^*(a|s)$ that maximizes the expectation of accumulated rewards $\mathbb{E}[\mathcal{R}_t]$ in \eqref{eq:ReturnRt}. The maximization problem can be resolved using diverse methods. In this paper, we adopt the well-known proximal policy optimization (PPO) algorithm \cite{schulman2017proximal}. Assume that the policy $\pi(a|s)$ is approximated using a deep neural network with the parameter set $\theta$, so the parameterized policy is denoted by $\pi_\theta(a|s)$. In PPO, the policy parameter is optimized by maximizing a surrogate objective as below. to limit the update of parameters into the trust region as follows \cite{schulman2017proximal}.
 \begin{equation}  \label{eq:SurrObj}
 \begin{array}{ll}
      L^{CLIP} =& \mathbb{E}\left[\min\left( \frac{\pi_\theta(a_t|s_t)}{\pi_{\theta_{old}}(a_t|s_t)}A_t, \right.\right.\\
      & \left.\left.{\rm clip}(\frac{\pi_\theta(a_t|s_t)}{\pi_{\theta_{old}}(a_t|s_t)}, 1-\epsilon, 1+\epsilon)A_t\right)\right]
 \end{array}
 \end{equation}
 where $A_t$ is the advantage function and $\epsilon$ is the clip parameter. More details can be found in \cite{schulman2017proximal}.

\subsection{Artificial Potential Field}\label{sec:APF}
Artificial potential field methods guide robots to their predefined goals by the combination of multiple forces. The attractive force exerted on robot $i$ is defined as follows.
	\begin{equation}
	\boldsymbol{F}_{a,i} = \frac{\boldsymbol{p}_{g,i} - \boldsymbol{p}_i}{d_{g,i}} \label{eq:AttrForce}
	\end{equation}
Accordingly, the repulsive force from an obstacle is
		\begin{equation}
            \label{eq:repulsive_force}
		\boldsymbol{F}_{r,i} = \left\{
		\begin{aligned}
		&\eta\left( \frac{1}{d_{o,i}} - \frac{1}{\rho} \right) \frac{\boldsymbol{p}_i - \boldsymbol{p}_{o,i}}{d_{o,i}^3}&\text{if } d_{o,i} \leq \rho\\
		&0 &\text{if } d_{o,i} > \rho
		\end{aligned}
		\right.
		\end{equation} \label{eq:RepForce}
 where $\eta$ is the scale factor. $\boldsymbol{p}_{o,i}$ is the position of the nearest obstacle to robot $i$. $\rho$ is the influence range of obstacles.
 
The inter-robot force that keeps robots away from each other is  
 \begin{equation} \label{eq:individual_force}
	\boldsymbol{F}_{in,i} = \sum_{j\neq i, j \in \mathcal{N}_i} \left( 0.5 - \frac{\lambda}{d_{j,i}} \right) \frac{\boldsymbol{p}_j-\boldsymbol{p}_i}{d_{j,i}}
	\end{equation}
where $\mathcal{N}_i$ is the set of robots that can be detected by robot $i$, namely $\mathcal{N}_i=\left\{j\vert\; \forall\;j\in\mathcal{V}\text{, }j\neq i\text{, \& }d_{j,i}<d_r\right\}$, $\lambda$ is a parameter adjusting the compactness of the multi-robot system. The smaller $\lambda$ is, the more 
compact the system is \cite{ZhangZheng2022}. 
The resultant force, therefore, is 
\begin{equation}
	\boldsymbol{F}_i = \boldsymbol{F}_{a,i} + \boldsymbol{F}_{r,i} + \boldsymbol{F}_{in,i}
\end{equation}
The direction of $\boldsymbol{F}_i$ is used as the commanded heading angle for robot $i$.
\section{Algorithm}\label{sec:alg}
    \begin{figure*}[t]
    \centering
		\includegraphics[width=0.95\linewidth]{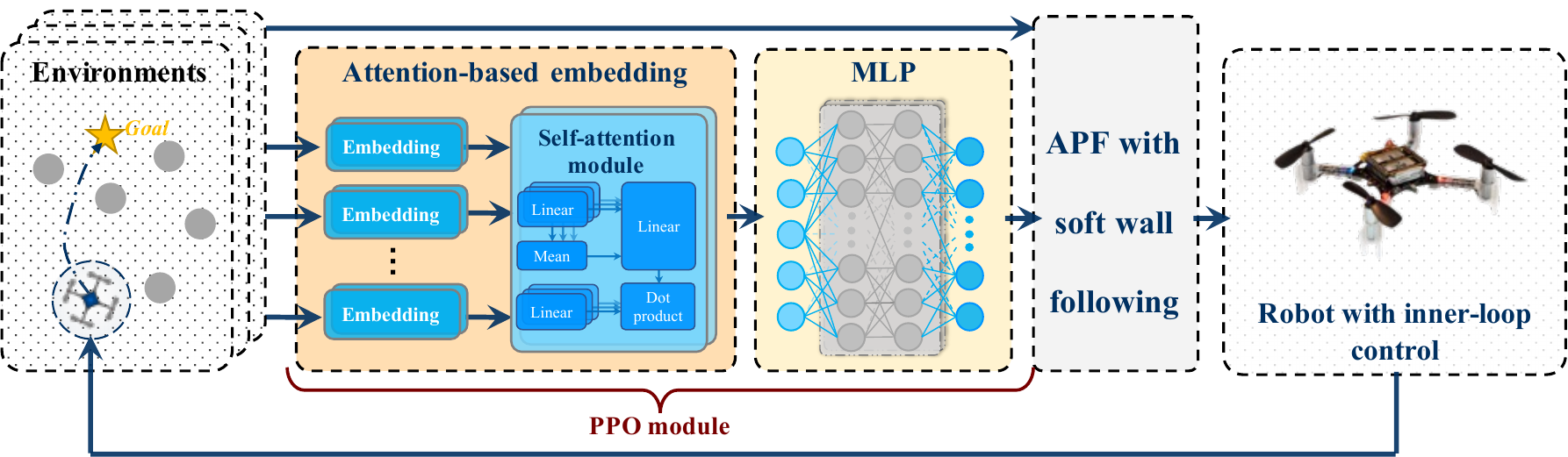}
		\caption{The overview of RPF. Observations are firstly transformed to a fixed-length high-dimensional vector by an attention-based embedding. It is then fed into the Multiple-Layer Perceptron (MLP) network. The whole network---attention-based embedding plus MLP---will be optimized by using a PPO agent. Hence, the PPO agent is used to encode environment information that will be used later to regulate a modified APF with a soft wall following mechanism.}
		\label{Fig:Framework}
    \end{figure*}

The reinforced potential field (RPF) algorithm proposed in this paper integrates attention-based embedding, PPO, and APF. Thereinto, the attention-based embedding transforms the observations into fixed-length vector according to the importance of neighboring robots and obstacles to the current decision-making process. Outputs of the attention-based embedding are fed into Multiple-Layer Perceptron (MLP) network to generate reasonable choices for a modified APF with a soft wall following mechanism. Therefore, we will describe the proposed algorithm from the aforementioned three aspects in this section. 
    
\subsection{Attention-based Embedding}
In fully-connected neural networks, the input size is required to be fixed. However, in a multi-robot scenario, the number of neighbours of a robot $i$ would be dynamically changing, so it would be inefficient to use a fully-connected neural network to encode surrounding environments. To tackle this problem, the mean embedding methods are employed by averaging all the information of observed neighbors \cite{huttenrauch2019deep}. However, it is obvious that the information from different neighbors contributes differently to the current decision-making process in a sophisticated multi-robot systems, \emph{e.g.}, gregarious animals. However, the pure average in the  mean embedding methods may underrate the information of a significant neighbor. Therefore, we introduce the self-attention mechanism into the observation embedding process in this paper to explicitly characterize the relative importance of surrounding observations, namely robot-robot interactions and robot-environment interactions.

The observations of each robot is firstly split to two parts, namely $\boldsymbol{o}_{i}=\left\{\boldsymbol{o}_{loc,i},\boldsymbol{o}_{nei,i}\right\}$. The first part $\boldsymbol{o}_{loc,i}$ includes the relative distance and azimuth angles of the nearest obstacle and the goal information, so $\boldsymbol{o}_{loc,i}=\left[d_{o,i}, \phi_{o,i},d_{g,i},\phi_{g,i}\right]^T$. The second part $\boldsymbol{o}_{nei,i}$ consists of the information of observed neighbors $\boldsymbol{o}_{nei,i}=\left\{\boldsymbol{w}_j|\forall\; j \in \mathcal{N}_i\right\}$, where $\boldsymbol{w}_j$ is
\begin{equation}\label{eq:NeighInf}
    \boldsymbol{w}_j = \left[d_{j,i},\phi_{j,i}, \psi_{j, i}\right]^T
\end{equation}
	

The attention-based embedding then employs two one-layer fully-connected networks to encode $\boldsymbol{w}_j$ into two fixed-length high-dimensional vectors $\boldsymbol{e}_j$ and $\boldsymbol{h}_j$, respectively. Both networks use ReLU activation functions.
    \begin{equation}
	\begin{aligned}
	\boldsymbol{e}_j &= \phi_e(\boldsymbol{o}_{loc,i}, \boldsymbol{w}_j) \\
	\boldsymbol{h}_j &= \Psi_h(e_j)
	\end{aligned}
    \end{equation}
The attention score is calculated by inputting $\boldsymbol{e}_j$ and the mean embedding of all observed neighbors $\boldsymbol{e}_m$ into another one-layer fully-connected network. This network also uses ReLU activation functions.
    \begin{equation}
	\begin{aligned}
	\boldsymbol{b}_j &= \Psi_b (\boldsymbol{e}_j, \boldsymbol{e}_m)
	\end{aligned}
    \end{equation}
where $\boldsymbol{e}_m = \frac{1}{\left| \mathcal{N}_i\right|}\sum_{j \in \mathcal{N}_i}\boldsymbol{e_j}$, where $\left| \mathcal{N}_i\right|$ is the number of observed neighbors by a robot $i$. 
The information of each observed neighbor is weighted according to their corresponding attention scores.
    \begin{equation}
	\boldsymbol{c}_i = \sum_{j \in \mathcal{N}_i}{{\rm softmax}(\boldsymbol{b}_j)\boldsymbol{h}_j}
    \end{equation}
Finally,  $\boldsymbol{o}_{loc,i}$ is concatenated with $\boldsymbol{c}_i$ to represent the observations of robot $i$ with a fixed-length vector $\hat{\boldsymbol{o}}_i$ while highlighting the information from significant neighbors. 

In the conventional artificial potential field methods, the scale factor $\eta$ in (\ref{eq:repulsive_force}) determines the strength of obstacles' influence, while $\lambda$ in (\ref{eq:individual_force}) adjusts the magnitude of individual force exerted from neighbors. Once these parameters are determined, they are invariant throughout the whole task. However, it is more promising to dynamically adjust these parameters online because their optimal values depend on the environment states in one episode. To encode the dynamic environment information into $\eta$ and $\lambda$, the outputs of the attention-based embedding will be fed into an MLP network. The whole framework will be optimized using PPO.

\subsection{Optimization by PPO}

	
The action space has two dimensions corresponding to $\eta$ and $\lambda$, respectively. The range of candidate action pairs is $[0,0.1]\times[0,5]$, which is empirically selected according to the specific scenarios used in Section \ref{sec:exp}. All robots share the same policy network, so the same reward function is used for all robots.  
The total reward function for each robot is defined as 
{
\begin{equation}
	R = R_{m} + R_{s} + R_{o} + R_{p}
	\end{equation}
 }
The reward $R_m$ encourages robots to reach their goals via the shortest trajectory. For a robot $i$, one has
    \begin{equation}
    R_m = \left\{
    \begin{aligned}
        &300 - 100\frac{d_{a}}{d_{s}}  &\text{if }d_{g,i}<r \\
        &0 &\text{otherwise}
    \end{aligned}
    \right.
    \end{equation}
where $d_{s}$ is the Euclidean distance between the start point and the goal position, and $d_{a}$ denotes the trajectory length achieved by the robot $i$. 

The reward function $R_{s}$ encourages robots to move smoothly by giving a punishment of $-5$ when the difference of headings at two adjacent timesteps exceeds $45^\circ$. 


The reward function { $R_o$ } is used to prevent robots from collisions with obstacles. For a robot $i$, one has 
	\begin{equation}
	R_o = \left\{
	\begin{aligned}
	-100,\quad &{\rm if }\  d_{o,i} < r\\
	-20,\quad &{\rm if }\  r \leq d_{o,i} < 2r\\
	0,\quad &{\rm otherwise}
	\end{aligned}
	\right.
	\end{equation}
The reward $R_p$ provides a dense reward to accelerate the learning process. For a robot $i$, one has 
    \begin{equation}
        R_p = \left\{
	\begin{aligned}
	1 - \frac{d_{g,i}}{d_m},\quad &{\rm if}\ d_{g,i} < d_m \\
	0,\quad &{\rm otherwise}
	\end{aligned}
	\right.
    \end{equation}
where $d_{g,i}$ is the relative distance of the goal, and $d_m>0$ is a hyperparameter. 

To extend the classical PPO algorithm to multi-robot settings, the parameter sharing technique is employed. That is, all robots share the same policy network. The shared policy network updates parameters using transitions collected by all robots. The overall procedure of the proposed algorithm is shown in Algorithm \ref{Alg:1}.  

	\begin{algorithm}[htbp]
		\caption{Multi-robot Reinforced Potential Field}
		\label {Alg:1}
		\begin{algorithmic}[1]
         \State {Algorithm initialization}
			\For {episode = $1,2,\cdots,$}
			\State {Reset the environment}
			\For {$t=1,2,\cdots,$}
			\For{$i = 1,2,\cdots, N$ }
			\If {$d_{g,i}>r$ }
			\State {Obtain $(\eta,\lambda)$ from $\pi_\theta(a|s)$}
			\State {Calculate $\boldsymbol{F}_{a,i},\boldsymbol{F}_{r,i},\boldsymbol{F}_{in,i}$ by APF}
            \If {$\boldsymbol{F}_{ar,i}^T\boldsymbol{F}_{a,i} < 0$}
            \State {Choose $n_{1,2}$ as $\boldsymbol{F}_i$}
            \Else \If {$\boldsymbol{F}_{r,i}^T \boldsymbol{F}_{a,i} < 0$}
            \State {Calculate $\boldsymbol{F}_{soft,i}$ as $\boldsymbol{F}_i$}
            \EndIf
            \EndIf
		\State {Move along the direction of $\boldsymbol{F}_i$}
		\State {Get the reward $R$ and observation $\boldsymbol{o}_i$}
		\EndIf
		\EndFor
            
             \If {$t \ {\mathrm{mod} \ Z} = 0$}
		\State {Update the PPO agent for $K$ epochs}
		\EndIf
            \EndFor
		\EndFor
		\end{algorithmic}
	\end{algorithm}

    \subsection{APF with soft wall following}
\begin{figure}[htbp]
        \includegraphics[width=\linewidth]{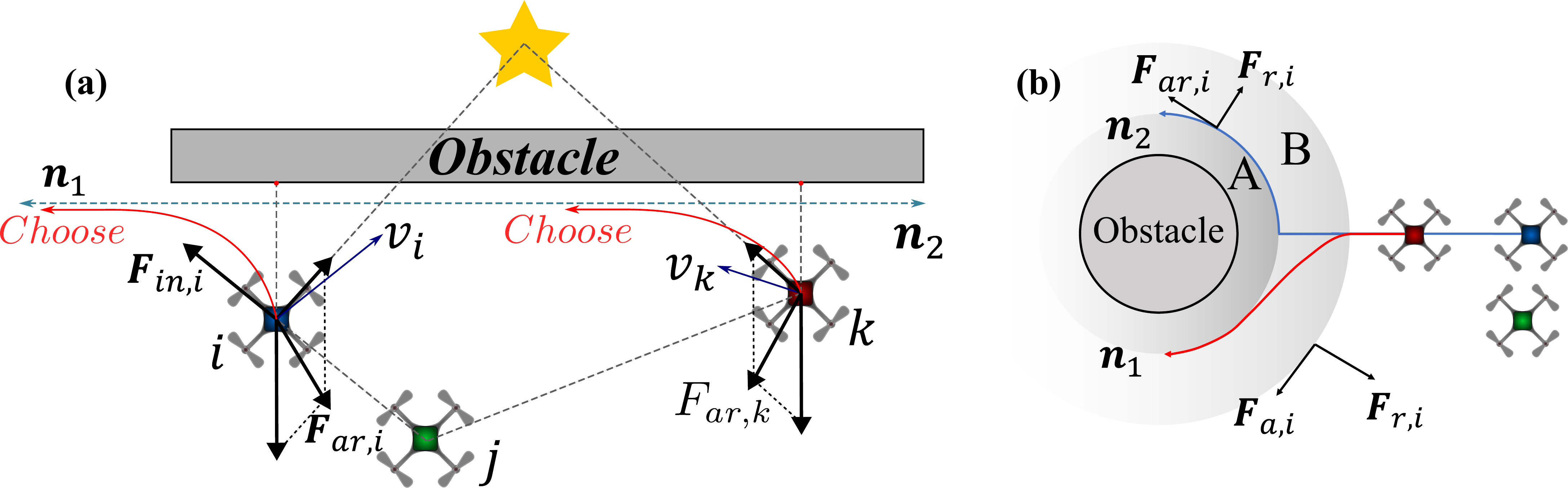}
		
        \caption{Wall Following Rule is shown in (a): agent $i$ (blue), agent $j$ (green), agent $k$ (red) are moving around an obstacle. The soft extension of wall following is shown in (b). In area A, the angle between the resultant force $F_{ar,i}$ and the attractive force $F_{a,i}$ exceeds 90°, where wall following rule is active. In Area B, the angle between the attractive $F_{a,i}$ and the repulsive $F_{r,i}$ exceeds 90°, where robots move according to soft wall following rule.}
        \label{fig:wall-following}
    \end{figure}

    Conventional artificial potential field method mentioned in Sec \ref{sec:APF} {can} be stuck in a location, especially when $\boldsymbol{F}_{r,i} = \boldsymbol{F}_{a,i}$ and $\boldsymbol{F}_{in,i}=0$. In order to tackle this problem, our previous work introduced a wall following rule \cite{ZhangZheng2022}. The wall-following rule means that when the agent is close to the obstacle, it should move along the edge of obstacles. As shown in Fig \ref{fig:wall-following}, $\boldsymbol{n}_1$ or $\boldsymbol{n}_2$ is the better direction for robots rather than $\boldsymbol{F}_{ar, i} =\boldsymbol{F}_{a,i}+\boldsymbol{F}_{r,i}$.
    The wall following rule is activated when the angle between $F_{ar,i}$ and  $F_{a,i}$ exceeds $90^{\circ}$. 
    
    The selection of $\boldsymbol{n}_1$ and $\boldsymbol{n}_2$ depends on the current heading direction and inter-robot force $\boldsymbol{F}_{in,i}$.
    If $\boldsymbol{F}_{in,i}$ exceeds the specified a threshold $\bar{F}_{in,i}$, we select from $\boldsymbol{n}_1$ and $\boldsymbol{n}_2$ the one that has a smaller angle with $\boldsymbol{F}_{in,i}$ as shown in Fig. \ref{fig:wall-following} (a). Otherwise, choose the one with a smaller angle with the current motion direction as the robot $k$ in Fig. \ref{fig:wall-following}.

When the angle between $\boldsymbol{F}_{ar,i}$ and $\boldsymbol{n}_{1}$ (or $\boldsymbol{n}_{2}$) is very large, there might be a sharp turn as the blue trajectory shown in Fig. \ref{fig:wall-following}(b). To solve the problem, we introduce a soft rule to the wall following method. The nearby external area of an obstacle is divided into two sub-areas as shown in Fig. \ref{fig:wall-following}(b). In sub-area A, a robot $i$ chooses $\boldsymbol{n}_1$ or $\boldsymbol{n}_2$ according to the aforementioned wall following rule.  In the sub-area B, the robot chooses a direction $\boldsymbol{F}_{soft}$ between $\boldsymbol{n}_{1\text{ or }2}$ and $\boldsymbol{F}_{ar,i}$. Hence, $\boldsymbol{F}_{soft,i}$ is defined as
    \begin{equation}
	\boldsymbol{F}_{soft,i} = \frac{\boldsymbol{F}_{ar,i} + 2\|\boldsymbol{F}_{r,i}\|  \boldsymbol{n}_{1\text{ or }2}}{\|\boldsymbol{F}_{ar,i} + 2\|\boldsymbol{F}_{r,i}\|  \boldsymbol{n}_{1\text{ or }2}\|}
    \end{equation}
    
    At the exact moment that a robot $i$ enters the sub-area B, $\boldsymbol{F}_{soft,i}$ is the same as $\boldsymbol{F}_{ar,i}$. When agents get closer to obstacle, $\boldsymbol{F}_{soft,i}$ becomes closer to $\boldsymbol{n}_{1\text{ or }2}$.

    The sub-area A and B around an obstacle in Fig \ref{fig:wall-following} are defined as follows: If $\boldsymbol{F}_{ar,i} ^T \boldsymbol{F}_{a,i} < 0$, it belongs to sub-area A; if $\boldsymbol{F}_{r,i}^T \boldsymbol{F}_{a,i} < 0$, it is in sub-area B.
    \begin{equation}
	\boldsymbol{F} = \left\{
	\begin{aligned}
	&\boldsymbol{n}_{1\text{ or }2}, &\boldsymbol{F}_{ar,i} \cdot \boldsymbol{F}_{a,i} < 0 \\
	&\boldsymbol{F}_{soft}, &\boldsymbol{F}_{r,i}^T \boldsymbol{F}_{a,i} < 0\  and \boldsymbol{F}_{ar,i} \cdot \mathbf{F}_{a,i} \geq 0
	\end{aligned}
	\right.
    \end{equation}

 \begin{figure}[htbp]
     \centering
    \includegraphics[width=\linewidth]{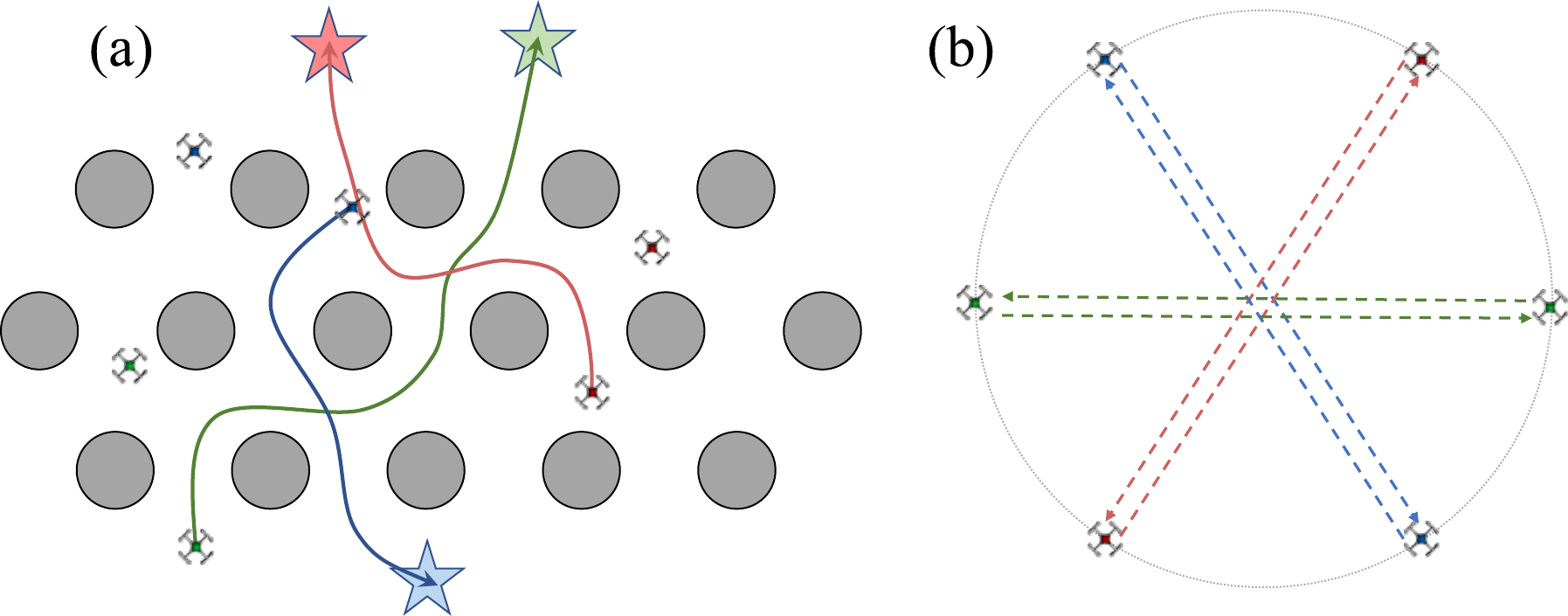}
     \caption{Arena for algorithm training. (a) Six robots go to their destinations, while avoiding dense obstacles (gray). The start point and goal positions of robots are generated randomly. (b) Six robots are distributed on a circle with random initial positions. Robots swap their locations with another one on the other side.}
     \label{fig:train_scene}
 \end{figure}

 \begin{table}[htbp]
     \centering
     \caption{Traning setup}
     \label{tab:train_params}
     \begin{tabular}{c|c}
          \bottomrule
          Parameters & Values \\
          \midrule
          Initial learning rate $\alpha_0$ & 0.0003 \\
          $\beta$ & 0.999 \\
          Actor neural network &  MLP with 2 hidden layers \\
          &(256 neurons per hidden layer) \\
          Critic neural network &  MLP with 2 hidden layers \\
          &(256 neurons per hidden layer) \\
          Sample batch size $Z$ & 100 \\
          PPO epochs $K$ & 1 \\
          Training episodes & 1000 \\
          Maximum steps per episode & 1000 \\
          $\tau$ & 0.9 \\
          $\gamma$ & 0.999 \\
          Clip parameter $\epsilon$ & 0.2 \\
          $c_1$ & 0.5 \\
          $c_2$ & 0.001 \\
          \bottomrule
     \end{tabular}
 \end{table}
	
	\section{Simulation and experiments}
 \label{sec:exp}
 In this section, numerical simulation and experimental results are presented. The proposed algorithm is trained in two different arenas with six robots as shown in Fig. \ref{fig:train_scene}. The trained results are thereafter evaluated in both numerical simulations with different setups and real-world experiments. 
 
 In the first arena, the environment is occupied with many obstacles as shown in Fig. \ref{fig:train_scene} (a). The obstacles have the same radius of $0.5$ $m$. Both the initial and goal positions of each robot are randomly given. All robots need to reach their goal positions, while avoiding collisions with each other and obstacles. The second training arena is shown in Figure \ref{fig:train_scene} (b). It is an open environment with no obstacles. All robots need to avoid collisions with each other. Initially, all robots are distributed evenly on a circle with a radius of $2$ $m$. The initial positions of all robots are randomly generated. Robots are required to swap their locations.  

 The same setup is chosen for the two aforementioned training arenas. Some training details are summarized in Table \ref{tab:train_params}. In the training, the desired speed of agents is set to $0.5$ $m/s$. The safety radius $r$ of robots is $0.1$ $m$. The perception range $d_p$ of a robot is chosen to be $6$ $m$. The time step at training is set to be $0.1$ $s$. The hyperparameter $d_m$ is chosen to be $10$ $m$. The influence range $\rho$ of an obstacle is chosen $10$ $m$. The wall following threshold $\bar{F}_{in,i}$ is picked to be $1$, $\forall$ $i\in\mathcal{V}$. The learning rate decays as follows. 
\begin{equation}
        \alpha = \alpha_0 \times \beta ^ {j}
\end{equation}
where $\alpha_0$ is the initial learning rate,  $\beta$ is a hyperparameter, and $j$ is the number of the current episode. The values of $\alpha_0$, $\beta$ are given in Table \ref{tab:train_params}. 

The proposed RPF is compared with three different methods, namely RPF without attention, vanilla PPO, and vanilla APF. In the comparison, the proposed RPF in this paper is termed as ``RPF 1''. When the attention module is replaced by pure linear embedding as in our previous work \cite{ZhangZheng2022}, we use the term ``RPF 2'' to represent it. The ``RPF 2'' algorithm is trained based on exactly the same reward functions with ``RPF 1'' using the same training setup. 

For vanilla PPO, we found that it's very difficult to train a policy in the cluttered environment shown in Fig \ref{fig:train_scene} (a). For comparison, we train the vanilla PPO policy in an environment with smaller obstacles. The training environment would be easier than the one used in ``RPF 1'' and ``RPF 2''. Hence,  the radii of all obstacles are set to be $0.1$ $m$. The locations of the obstacles are the same with those in Fig \ref{fig:train_scene} (a). Other learning configurations are the same  with those as ``RPF 1''  as given in table \ref{tab:train_params}. The decision of PPO robot is generated as follows.
\begin{equation}
    \boldsymbol{F}_i = \frac{\boldsymbol{v}_i + a_t \boldsymbol{v}_{i\perp}}{\|\boldsymbol{v}_i + a_t \boldsymbol{v}_{i\perp}\|}
\end{equation}
where $\boldsymbol{v}_{i\perp}$ is a vector perpendicular to $\boldsymbol{v}_i$, and $a_t$ is the output by the PPO policy with $a_t\in[-2.5, 2.5]$.

We choose the parameters of a vanilla APF to be $\eta=0.05$ and $\lambda=2$.  For a fair comparison, we choose two metrics to evaluate the performance of all methods in the simulation. 

The \textbf{traveling distance} metric $l_i$ evaluates the average total traveling length by a robot, which is defined as below.
 \begin{equation}
   l_i = \sum_{i \in \mathcal{V}}\sum_{t}^T ||\boldsymbol{v}_i(t)||\Delta t
   \label{eq:distance}
 \end{equation}
  where $T$ represents the total timesteps for robots to reach their goal. The large the traveling distance metric is, the poor the algorithm's performance is. 

The \textbf{motion smoothness} metric $\xi_i$ evaluates how oscillatory the motion of a robot, which is defined as 
    \begin{equation}
        \xi = \frac{\sum_{i\in\mathcal{V}} \sum_t^T ||\Delta \boldsymbol{v_i}|| / ||\boldsymbol{v_i}||}{T}
        \label{eq:change_angle}
    \end{equation}
Similarly, the large the  motion smoothness metric is, the poor the performance of an algorithm is. 

 \subsection{Simulation Evaluation} \label{subsec:sim}
\begin{figure}[htbp]
     \centering
     \includegraphics[width=\linewidth]{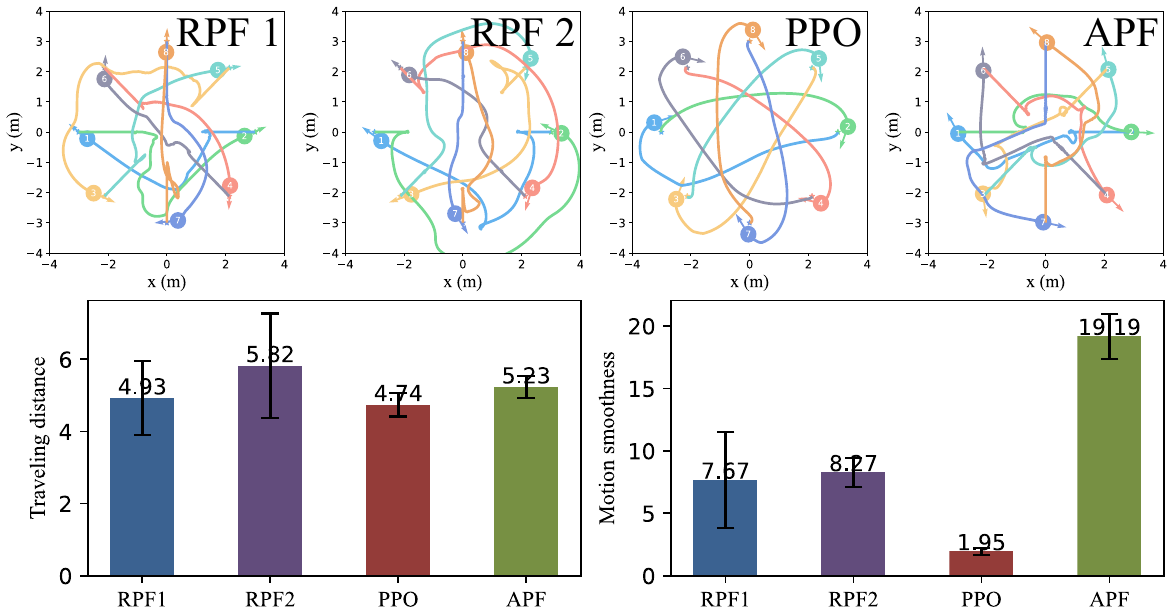}
     \caption{Simulation evaluation in an open environment. \textbf{RPF 1}: Reinforced Potential Field; \textbf{RPF 2} without the attention mechanism. All $8$ robots are initially distributed on a circle with a radius of $3$ m and swap their locations.
     }
     \label{fig:compare_circle}
 \end{figure}

\begin{figure}[htbp]
    \centering
    \includegraphics[width=0.98\linewidth]{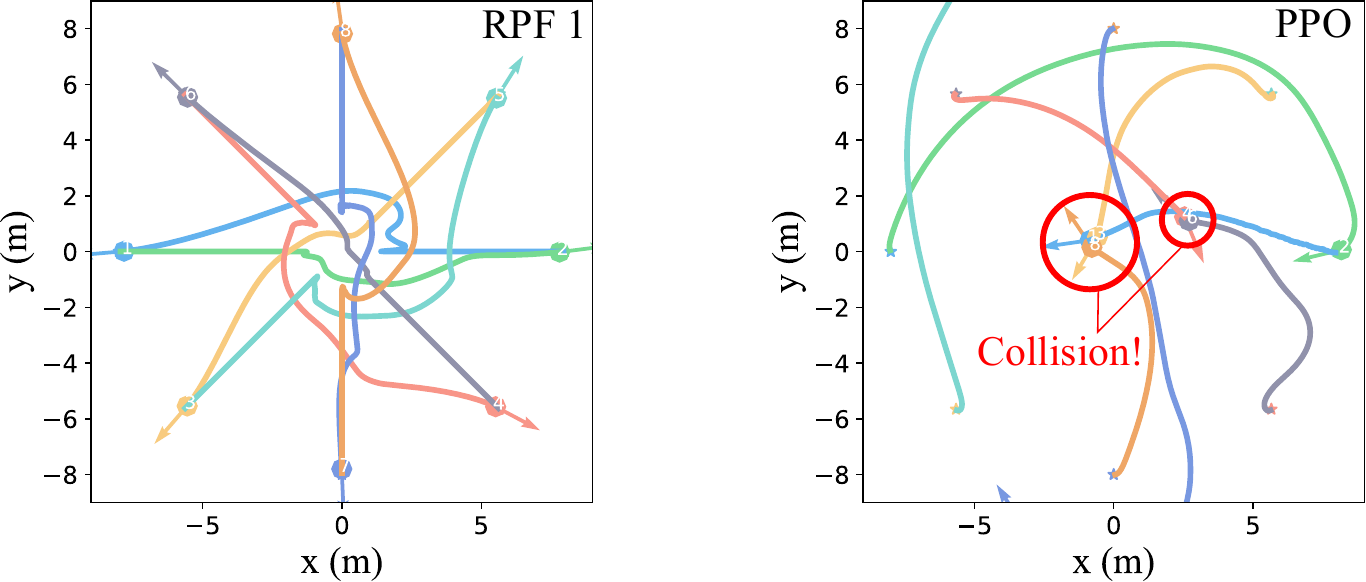} 
    \caption{Comparison of RPF 1 and PPO with $8$ robots initially distributed on a circle with a radius of $8$ $m$.
    }
    \label{fig:ppo_fail}
 \end{figure}

\begin{figure}[htbp]
     \centering
     \includegraphics[width=0.98\linewidth]{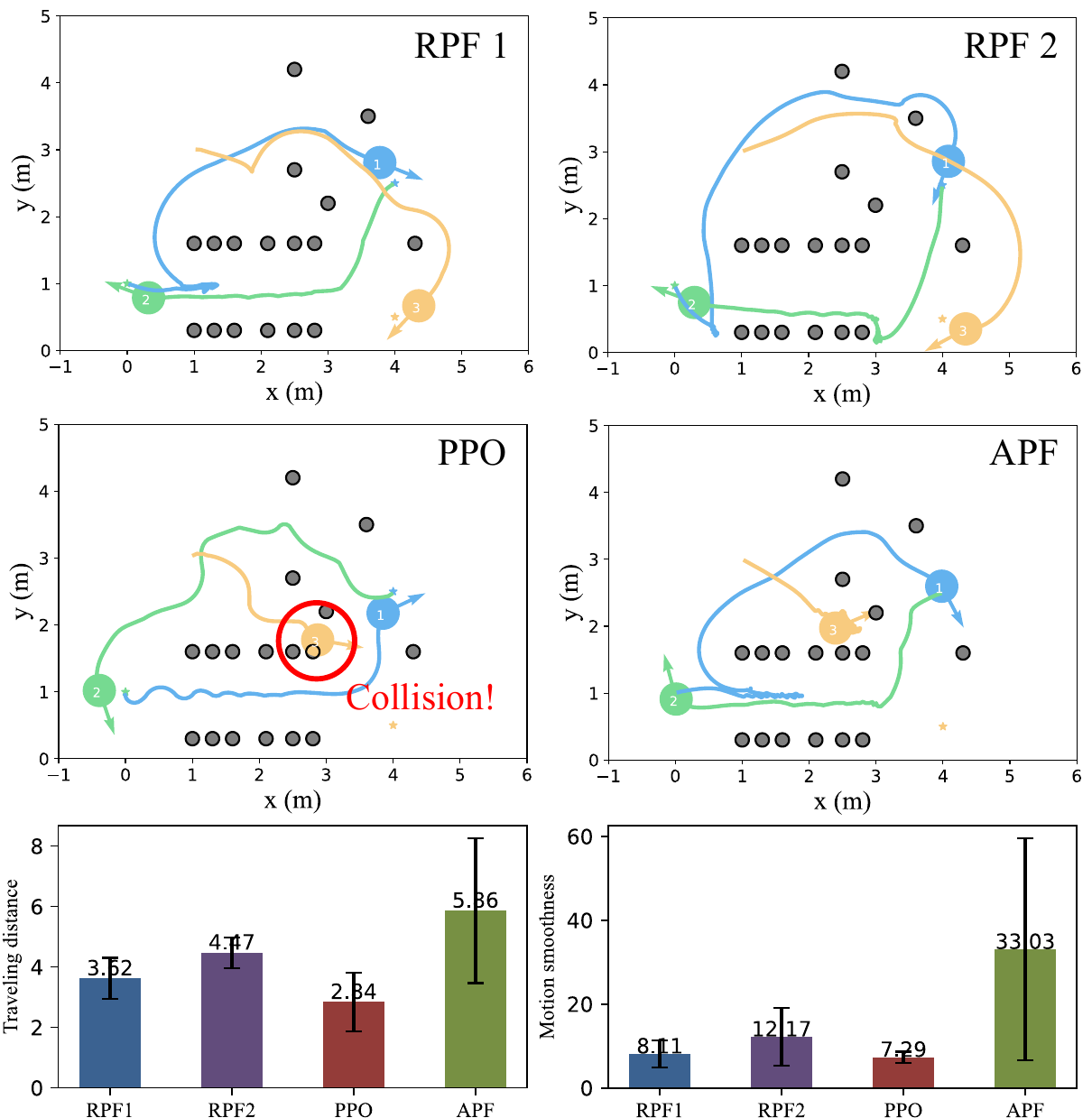}
     \caption{Simulation evaluation in an obstacle-rich environment. \textbf{RPF 1}: with attention; \textbf{RPF 2}: without  attention. Three robots are initially distributed at different locations in an environment different from the training setup, such as the obstacle distribution and obstacle size ($0.1$ 
 $m$ in the evaluation and $0.5$ $m$ in the training).
 }
     \label{fig:compare_lab}
 \end{figure}

 In the simulation evaluation, the trained proposed algorithm is compared with the other three algorithms in three different scenarios to show its efficiency and generalization.  In the first two scenarios, open environments with no obstacles are considered. In the second scenario, we built up a new evaluation arena with different obstacle distributions. All simulations and comparisons are given in Fig. \ref{fig:compare_circle} - \ref{fig:compare_lab}. Both the traveling distance and motion smoothness metrics in Figs. \ref{fig:compare_circle} and \ref{fig:compare_lab} are averaged among all robots to evaluate the whole performance of the algorithms in a multi-robot setting.

 In the first evaluation, we consider $8$ robots distributed on a circle with a radius of $3$ $m$ and swap locations with robots on the other side. This scenario is very similar to our second training arena but with a different number of robots and a different size of circle. The robot trajectories are shown in Fig. \ref{fig:compare_circle}. In this scenario, PPO has the best performance in by both the traveling distance metric and the motion smoothness metric as shown in Fig. \ref{fig:compare_circle}. The APF method has the worst performance by both metrics. The $RPF$ 1 (with the self-attention module) performs better than $RPF$ 1 (with no self-attention). It is that it is necessary to rank the importance of surrounding information. 

 Although the vanilla PPO has the best performance in the first scenario, it cannot generalize well to a scenario with more difference as shown in Figs. \ref{fig:ppo_fail} and \ref{fig:compare_lab}. In the second evaluation in Figs. \ref{fig:ppo_fail}, we further compare our method $RPF$ 1 with the vanilla PPO in a different scenaro. In this case, $8$ robots are initially distributed on a circle with a radius of $8$ $m$ that is large than the detection range ($6$ $m$) of a robot. The same position swap task is performed. In such a scenario, the vanilla PPO failed to complete the task safely.

 In the third scenario, we further evaluate the performance of all four algorithms in a different environment. In this scenario, the second robot collides with an obstacle, thus leading to a mission failure. Due to the mission failure of the second robot, the PPO algorithm has small values by both the distance metric and the motion smooth metric. In comparison with PPO, both $RPF$ 1 and $RPF$ 2 show better generalization, but  $RPF$ 1 has much improvement by both the traveling distance and motion smoothness metrics. It further demonstrates the efficiency of the introduced attention-based observation embeddings. 
 

 \subsection{Experiment Evaluation}
 \begin{figure}[htbp]
     \centering
     \includegraphics[width=\linewidth]{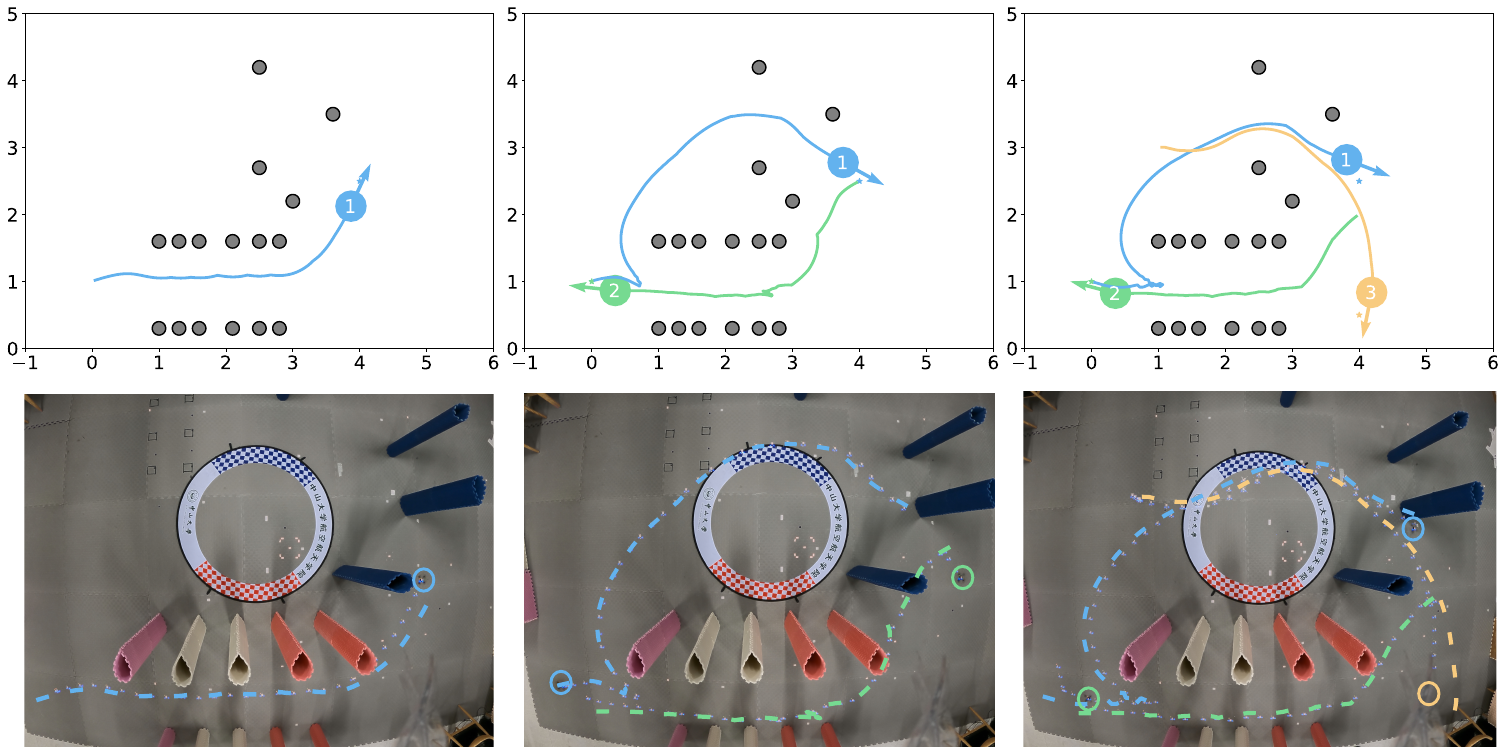}
     \caption{{Three experiments are conducted.} \textbf{Top}: Experiment data replay; \textbf{Bottom}: Snapshots of experiments.}
     \label{fig:real_exp}
 \end{figure}
 Experiment evaluation is performed to demonstrate the competence of the proposed algorithm in a real-life platform. A similar environment to the third simulation in Section \ref{subsec:sim} as shown in Fig \ref{fig:compare_lab}. In the experiment, we apply the RPF 1 algorithm to Crazyflie quadrotors without any modification. The OptiTrack optical motion capture system is used to measure the states of quadrotors. The Crazyflie drones can track position commands from the high-level planner, using the implemented inner-loop controller. 

 Three different experiments are performed with a different number of robots as shown in Fig \ref{fig:real_exp}. A robot would change its trajectory if more robots are considered in the environment. The experiment also illustrates the generalization of the proposed algorithm to real-life systems. 
 

\section{Conclusion} \label{sec:Conclusion}
    Reinforced potential field, a novel motion planning algorithm integrating deep RL with APF, was presented in this paper. The proposed design was able to make active responses to dynamic changes in surrounding environments. An observation embedding with self-attention mechanism was developed to implicitly environment dynamic changes. A soft wall-following rule was presented to further improve the motion smoothness performance.
    The performance improvement of our algorithm over existing benchmarks, including PPO and APF, was demonstrated via numerical simulations. Real experiments were also performed to show the competence of the proposed method in real systems. In future works, we aim to integrate real sensor observations into our method for implementation in more diverse scenarios.
	
	\bibliography{references.bib}
	\bibliographystyle{IEEEtran}
	
\end{document}